\documentclass{article}
\usepackage{ijcai19}
\usepackage{subfigure}
\usepackage{times}
\usepackage{soul}
\usepackage{url}
\usepackage[hidelinks]{hyperref}
\usepackage[utf8]{inputenc}
\usepackage[small]{caption}
\usepackage{graphicx}
\usepackage{amsmath}
\usepackage{booktabs}
\usepackage{algorithm}
\usepackage{algorithmic}
\usepackage{multirow}
\title{Listwise View Ranking for Image Cropping}
\author{
Weirui Lu$^1$\and
Xiaofen Xing$^1$\and
Bolun Cai$^1$\and
Xiangmin Xu$^1$\\
\affiliations
$^1$South China University of Technology, China\\
\emails
\{luweirui1022, caibolun\}@gmail.com,
\{xmxu,xfxing\}@scut.edu.cn,
}

\begin{document}

\maketitle

\begin{abstract}
Rank-based Learning with deep neural network has been widely used for image cropping. However, the performance of ranking-based methods is often poor and this is mainly due to two reasons: 1) image cropping is a listwise ranking task rather than pairwise comparison; 2) the rescaling caused by pooling layer and the deformation in view generation damage the performance of composition learning. In this paper, we develop a novel model to overcome these problems. To address the first problem, we formulate the image cropping as a listwise ranking problem to find the best view composition. For the second problem, a refined view sampling (called RoIRefine) is proposed to extract refined feature maps for candidate view generation. Given a series of candidate views, the proposed model learns the Top-1 probability distribution of views and picks up the best one. By integrating refined sampling and listwise ranking, the proposed network called LVRN achieves the state-of-the-art performance both in accuracy and speed.
\end{abstract}

\section{Introduction}

Image cropping is a common photo manipulation process, which improves the overall composition by removing unwanted regions.
Image cropping is widely used in photographic, film processing, graphic design, and printing businesses. Recent methods tend to learn photo composition and extract well-composed regions from ill-composed photo.

With the development of deep learning, most of researchers have devoted their efforts to proposing deep networks based on ranking approach. For ranking-based training, a number of candidate views in each image are labelled with the aesthetic ordering. Then, the image cropping task is formalized as classification of view pairs into two categories (correctly ranked and incorrectly ranked). Finally, sliding window \cite{Chen2017Learning}, detector \cite{wang2017deep} or reinforcer \cite{Li2018A2} is adopted to finding the best view.

In \cite{chen2017quantitative}, Chen \emph{et al.} first investigated learning-to-rank methods for image cropping. View finding network (VFN) \cite{Chen2017Learning} based on a pairwise ranking layer is proposed to model the photo composition and crop image by sliding window. Wei \emph{et al.} trained a view evaluation network (VEN) \cite{wei2018good} with the pairwise siamese architecture. Inspired by knowledge distillation \cite{hinton2015distilling} and anchor boxes \cite{liu2016ssd}, view proposal network (VPN) is proposed to transfer knowledge from VEN. In \cite{Li2018A2}, reinforcement learning is adopted to crop image step by step, and each step is controlled by the aesthetic score generated by a pre-trained VFN.

However, these ranking-based cropping methods are often poor in performance, which is mainly due to two reasons:

First, pairwise training is unsuitable for image cropping process. In image cropping, the main goal is to pick up the best composed view from a list of candidate views. That is, image cropping is a listwise ranking task rather than pairwise comparison. In addition, pairwise training heavily depends on careful pair selection, because the samples with various distribution will result in training bias. Therefore, pairwise training significantly increases the computational complexity and make the training procedure unstable.

Second, coarse feature extracted from convolutional neural network (CNN) will affect the accuracy of model learning. Previous methods crop and warp the views in raw images or feature maps, and then calculate the rank score for each one. Pixel-accuracy is important in image cropping rather than object classification, and it will be reduced by warp operation. In addition, rescaling caused by the pooling layers will reduce the sampling resolution and damage the composition learning.

To overcome these problems of image cropping, we propose a listwise ranking method with refined view sampling. In refined view sampling, a novel region of interest (RoI) operation called RoIRefine is proposed to extract refined feature maps of candidate views. Instead of carefully selecting the view pairs, we take advantage of all annotated views and train the model with a listwise ranking loss.

In summary, our main contributions are:
\begin{itemize}
\item We learn deep network for image cropping with listwise ranking.
\item We propose a refined view sampling named RoIRefine to alleviate the problem of rescaling and distortion.
\item The proposed model significantly outperforms the state-of-the-art methods in both accuracy and speed.
\end{itemize}

\section{Related Work}

\subsubsection{Image Cropping}
Image cropping is a common operation in image editing, which aims to find views with good photo composition.
A lot of methods have been proposed towards automating this task.
Previous cropping methods, in general, can be divided into attention-based or aesthetic-based approaches.
The attention-based methods focus on finding the most visually important area in the original image.
For example, ~\cite{marchesotti2009framework} trained a simple classifier on an annotated image database for generating attention maps.
In ~\cite{ciocca2007self}, visual saliency information, face and skin color detection results are combined for placing bounding box in image cropping.
For those aesthetic-based methods, they emphasize the general attractiveness of cropped image.
~\cite{fang2014automatic} proposed a aesthetic photo cropping system which combines three models: visual composition, boundary simplicity and content preservation.
A set of aesthetic quality classifiers were trained to discriminate the quality of candidate windows ~\cite{wang2017deep}.
With the development of datasets labelled by comparative aesthetic score, ranking-based methods are adopted to grade the composition of candidate windows~\cite{kong2016photo,chen2017quantitative}.
Recently, ranking-based methods together with other novel framework (e.g. knowledge transfer\cite{wei2018good} and reinforcement learning\cite{Li2018A2}) have achieved the state-of-the-art performance.

\subsubsection{Learning to Rank}
Ranking is widely used in information retrieval~\cite{yao2016deep}, recommender systems~\cite{li2016relaxed} and software engineering~\cite{xuan2014learning}.
In learning-to-rank task, training data consists of lists of items with some partial order which is specified between items in each list.
Most ranking algorithms are categorized into three groups by their input representation and loss function: the pointwise, pairwise, and listwise approach~\cite{liu2009learning}.

Pointwise approaches assume that each item in the training data has a numerical or ordinal score.
Then the learning-to-rank problem can be approximated by a regression problem.
Ordinal regression and classification algorithms can be used to predict the score of a single item.
For example, the perceptron ranking (PRank) algorithm was proposed to find a rank-prediction rule that assigns each instance a rank order ~\cite{crammer2002pranking}.

Pairwise approach formalizes the learning task as comparison of object pairs into two categories (correctly and incorrectly).
RankNet ~\cite{burges2005learning} learned a rank rule by using gradient descent methods and a natural probabilistic cost function on pairs of examples.
RankBoost ~\cite{freund2003efficient} used boosting to train ranking model by minimizing classification errors on instance pairs.

Listwise approaches try to directly optimize the value over all items on training data.
ListNet ~\cite{cao2007learning} tried to define a listwise loss function for learning to rank and introduces two probability models, respectively referred to as permutation probability and Top-1 probability.
Suppose that $\pi$ is a permutation on the n objects, and $\Phi\left(\cdot\right)$ is an increasing and strictly positive function.
Then, given the list of scores $s$, ListNet defines the probability of permutation $\pi$ as
\begin{equation}\label{eq:p_PI}
P_{s}(\pi) = \prod_{j=1}^n \frac{\Phi(s_\pi(j))}{\sum_{k=1}^n \Phi(s_\pi(k))}
\end{equation}
and the top one probability of object j is defined as
\begin{equation}\label{eq:top}
P_{s}(j) = \sum_{\pi(1) = j,\pi \in \Omega_n}P_{s}(\pi),
\end{equation}
where $\pi(1) = j$ means the $j$ object is ranked on top one in $\pi$ permutation. Thus, from Eq. \eqref{eq:p_PI} and  \eqref{eq:top}, we can obtain
\begin{equation}\label{eq:ps}
P_{s}(j) = \frac{\Phi(s_j)}{\sum_{k=1}^n \Phi(s_k)},
\end{equation}
where $s_j$ is the score of object $j = 1,2,\dots,n$.

In  general, with the use of top one probability, cross entropy is used to represent the distance between the two given score lists.

\section{The Proposed Approach}\label{sec:method}
\begin{figure*}[htb]
    \centering
    \includegraphics[width=1.0\linewidth]{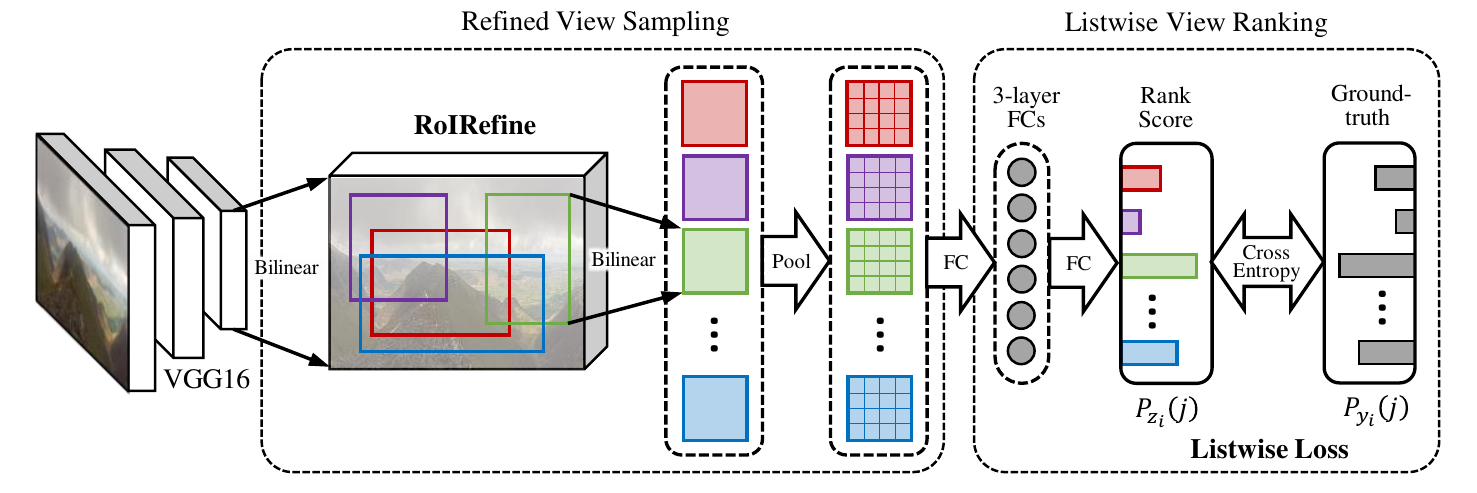}
    \caption{The framework of listwise view ranking for image cropping. The model first applies VGG16 (truncated before the last max pooling layer) as the backbone to extract the feature of input images. Given a series of candidate boxes, the refined view sampling integrates RoI features in each bounding boxes. And then we use 3-layer full connection (FC) to generate the ordering distribution, and apply cross entropy to measure the listwise distance.}
    \label{fig:framework}
\end{figure*}

In this paper, we propose a listwise view ranking network (LVRN) for image cropping. As illustrated in Figure \ref{fig:framework}, a refined view sampling (called RoIRefine) extracts high-resolution features to rank candidate views with listwise loss.

\subsection{Listwise View Ranking}

To address the shortcut of pairwise approaches, we formulate composition learning as a listwise ranking problem. In this paper, the proposed model listwisely ranks the candidate views and picks up the best one.

Given a set of annotation images $D = \{d_1,d_2,\dots,d_m\}$, each image $d_i$ consists of a list of candidate views $V_i = \{v_i^1, v_i^2, \dots, v_i^n\}$, where $m$ is the number of images and $n$ is the number of views. For each view $v_i^n$ in the $i$-th image, a rank score $y_i^n$ is labelled to represent the relative degree of view composition. For instance, the number of views $n$ is 24 in CPC dataset labelled with listwise protocol. In the view ranking network, we denote the rank function as $f\left(\cdot\right)$, which takes a view $v_i^j$ (sampled from image $d_i$) as input and then outputs a rank score $f(v_i^j)$. For the $i$-th image, we can obtain a list of scores $Z_i = \{f(v_i^1), f(v_i^2), \dots, f(v_i^n)\}$ from the list of views $V_i$. Therefore, the ranking function $f\left(\cdot\right)$ can be optimized by minimizing the loss between $Z_i$ and ground-truth scores $Y_i$.

Instead of pairwise approaches carefully selecting the training pairs, listwise learning removes the training bias as all candidate views are seen in each iteration. Even so, there are still a few view biases in the list -- the best composed view is important than the worse ones. To address this problem, a nonlinear transformation $\Phi\left(\cdot\right)$ is adopted to amplify the effect of the best one. We define $\Phi\left(\cdot\right)$ as a common increasing function:
\begin{equation}
\Phi\left(f(v_i^j)\right) = exp\left(f(v_i^j)\right)
\end{equation}
According to Eq. \eqref{eq:ps}, we rewrite the output scores to Top-1 probability as
\begin{equation}
P_{Z_i}(j) = \frac{\Phi(f(v_i^j))}{\sum_{k=1}^n \Phi(f(v_i^k))}.
\end{equation}
Similarly, the ground-truth score is rewrote as $P_{Y}$. Following \cite{cao2007learning}, we employ cross entropy as metric to minimize the distance between output probability $P_{Z}$ and ground-truth probability $P_{Y}$. The loss function is defined as
\begin{equation}
L(Y,Z) = -\sum_{i=1}^m \sum_{j=1}^n P_{Y_i}(j) log(P_{Z_i}(j)).
\end{equation}

The ranking function $f\left(\cdot\right)$ can be simply found by minimizing the loss function $L(Y,Z)$. Once the ranking function $f\left(\cdot\right)$ is learned, we simply use it to calculate the rank scores and crop the images from candidate views.

\subsection{Refined View Sampling}

Coarse features extracted from CNN backbone limit the performance of image cropping. Previous methods generate candidate views from images and then warp them to a fixed size (e.g., $227 \times 227$ in VFN ). However, warping is not suitable for composition learning and make the view deformed. The deformation of feature seriously damages the common composition rule, such as golden ratio, golden spiral and rule of thirds. In additional, the rescaling and multiple down-sampling in the CNN backbone make the model insensitive to view contents.

\begin{figure}[htb]
	\centering
    \subfigure[RoIPool: Crop+Pool]{\label{fig:roipool}\includegraphics[height=2.2cm]{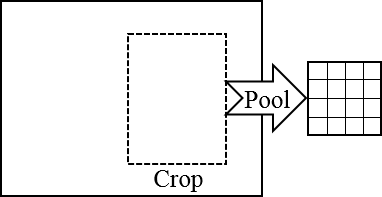}}~~
    \subfigure[RoIAlign: Interp+Crop+Pool]{\label{fig:roialign}\includegraphics[height=2.2cm]{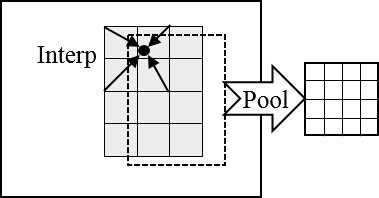}}
    \subfigure[RoIWarp: Crop+Interp+Pool]{\label{fig:roiwarp}\includegraphics[height=2.2cm]{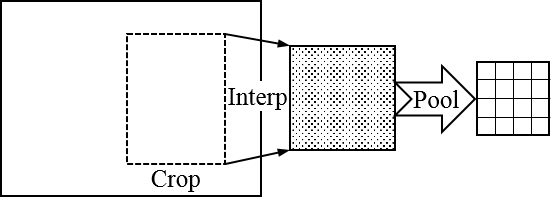}}
    \subfigure[RoIRefine: Interp+Crop+Interp+Pool]{\label{fig:roirefine}\includegraphics[height=2.2cm]{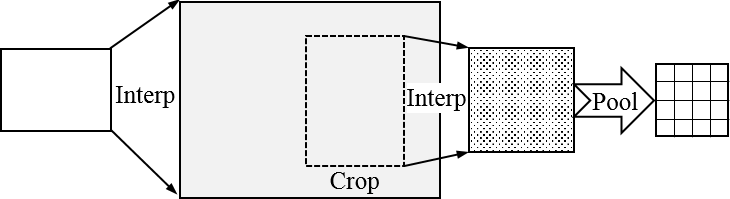}}
    \caption{Illustrations of RoI-aware operations and our RoIRefine.}
    \label{fig:roi}
\end{figure}

For view generation, there are three common RoI-aware operation shown in Figure \ref{fig:roi}:
\begin{itemize}
\item \textbf{RoIPool} \cite{girshick2015fast} (Figure \ref{fig:roipool}) is a standard operation for extracting a small feature map from each RoI. The quantized RoI is subdivided into spatial bins, and finally feature values covered by each bin are aggregated. The quantizations introduce misalignments between the RoI and the extracted features.
\item \textbf{RoIAlign} \cite{He2017Mask} (Figure \ref{fig:roialign}) removes the harsh quantization of RoIPool, properly aligning the extracted features. In each RoI, RoIAlign uses bilinear interpolation to compute the exact features at four regularly sampled locations, and aggregates the RoI features using max/average pooling.
\item \textbf{RoIWarp} (Figure \ref{fig:roiwarp}) operation is proposed in \cite{dai2016instance}. Unlike RoIAlign, RoIWarp crops a feature map region and warps it into a target size by interpolation. Even though RoIWarp also adopts bilinear resampling, it overlooks the alignment of floating-number RoI.
\end{itemize}

These RoI-aware operations are widely used in object detection and instance segment, but unsuitable for image cropping. Inspired by RoIAlign and RoIWarp, we propose an \textbf{RoIRefine} layer shown in Figure \ref{fig:roirefine} to extract high-quality features for reducing deformation. Our proposed change is simple: we sample the full-map features and resample the RoI-aware features to reduce deformation. The first bilinear interpolation improves the sampling resolution. Although the first bilinear interpolation does not increase additional information, the improvement of resolution makes the features sensitive to floating-number RoI. Without the first interpolation, we cannot achieve the float coordinate in the feature map, which means candidate boxes shift or rescale. In the other word, interpolation implements finer sampling with float-quantization. The second resampling avoids inconsistent between the feature maps and candidate views. In this paper, we simply upsample the full-map features to $2 \times$ size and resample the RoI-aware features to the size of $14\times14$. Considering the trade-off between performance and efficiency, $2 \times$ upsampling is the best choice as larger scale upsampling (4x or 8x) hardly improves performance. Compare to previous RoI-aware operations, RoIRefine leads to large improvements as shown in Section \ref{sec:roirefine}.

\subsection{Implementation}
In this paper, we initialize the backbone CNN with VGG16 pre-trained on ImageNet. All weights of the three FC layers are initialized with normal distribution (zero mean and 0.01 standard deviation), bias are set to zero and the channels are set to 1024, 512 and 1, respectively. The proposed model is trained on CPC dataset \cite{wei2018good} including 10,797 images, each with 24 candidate views. We directly rank 24 views and assign the order of views as ground-truth rank score.

During training, the images are resized to 224 $\times$ 224 regardless of its original size. Resizing the original image to fixed size is to fit the VGG-16 pretrained on 224x224 images (ImageNet), and is beneficial to model fine-tuning. Although resizing the original image does result global deformation, but its effect is weak. Global deformation does not affect listwise ranking because the ranking objects are views instead of original images. Every candidate views in one list only have the same global deformation, and listwise ranking loss is not sensitive to global deformation.

We trained the network for 10 epochs using stochastic gradient descent (SGD) with momentum of 0.9 and learning rate of 0.001 that decays by 0.1 after 4 epochs. The batch-size is set to 50 that means each mini-batch including $50\times24$ candidate views cropped from 50 images. Early stopping was adopted based on validation results on FCDB dataset \cite{chen2017quantitative}.

\section{Experiments}
We validate the effectiveness of the proposed model on two public image cropping databases (FCDB \cite{chen2017quantitative} and FLMS \cite{fang2014automatic}). We also compare the time efficiency on a GPU to existing image cropping models in Table \ref{tab:time}.

\subsection{Experimental Settings}

To evaluate our model, we utilize the sliding window strategy of \cite{Chen2017Learning} to generate candidate views and choose the views with best rank score $f\left({v_i^j}\right)$. Here we set the size of search windows among [0.6, 0.65, 0.7, ..., 0.9] and the aspect ratio among [1:1, 3:4, 4:3, 9:16, 16:9]. To refine candidate views, we adopt non-maximum suppression (NMS) based on overlap ratio between candidate views and original image to generate 1,745 candidate boxes.

\subsubsection{FCDB Dataset}
FCDB contains 348 test images and each image is labelled by a photography hobbyist. To evaluate the generalization ability of our model, we adopt the same metrics as previous works \cite{chen2017quantitative,wei2018good}, including \emph{intersection-over-union} (IoU) and \emph{boundary displacement} (Disp). The IoU can be computed as
\begin{equation}
IoU = \frac{{Area_i^{gt} \cap Area_i^{pred}}}{{Area_i^{gt} \cup Area_i^{pred}}},
\end{equation}
where $Area^{gt}_i$ and $Area^{pred}_i$ denote the area of the ground-truth and best-ranking crop view, respectively. Boundary displacement is given by
\begin{equation}
Disp = \sum\limits_{j = 1}^4 {\frac{{\parallel \hat{B}_i^j - B_i^j\parallel }}{4}} ,
\end{equation}
where $\hat{B}^{j}_i$ and $B^{j}_i$ denote the four corresponding edges between the ground-truth and best-ranking crop view, respectively.

\begin{table}[!tb]
\centering
\begin{tabular}{c|c|c}
\hline
Methods                             & Avg. IoU       & Avg. Disp \\
\hline
AesRankNet \cite{kong2016photo}     & 0.4843        & 0.1400   \\
RankSVM \cite{chen2017quantitative} & 0.6020        & 0.1060   \\
VFN+SW \cite{Chen2017Learning}      & 0.6328        & 0.0982   \\
A2-RL \cite{Li2018A2}               & 0.6633        & 0.0892   \\
VPN \cite{wei2018good}              & 0.6641        & 0.0858   \\
\hline
\textbf{LVRN (Ours)}                       & \textbf{0.7100}  & \textbf{0.0735} \\
\hline
\end{tabular}
\caption{Quantitative evaluation on FCDB dataset.}
\label{tab:fcdb}
\end{table}

\subsubsection{FLMS Dataset}
FLMS contains 500 test images and each image has 10 annotations from 10 different persons. The evaluation metric is a little different as it has more annotations for each image than FCDB. Following previous methods, Top-1 maximum IoU is chosen as the evaluation metric. Top-1 means to pick up the best cropping views to compute the result. We compute the IoU between the ground-truth and Top-1 views, and then choose the maximum IoU as final results.

\subsection{Quantitative Evaluation}

In this section, we study the cropping accuracy of our model with the state-of-the-art methods. We evaluate the performance on FCDB and FLMS dataset. VFN uses the ground truth window as the candidate views which leads to remarkable improvement, and VPN performs a post-processing by discarding small views to improve performance. For comparison fairness, the results (shown in Table \ref{tab:fcdb} and Table \ref{tab:flms}) are evaluated without ground-truth windows and post-processing as \cite{Li2018A2}.

\subsubsection{FCDB Dataset}
As shown in Table \ref{tab:fcdb}, we evaluate the cropping performance on FCDB dataset. Besides of the methods discussed above (VFN, A2-RL and VPN), we choose two other pairwise learning-to-rank methods as baselines. AesRankNet \cite{kong2016photo} is proposed to rank photo aesthetics modelled by a pairwise loss function. RankSVM \cite{chen2017quantitative} uses AlexNet to extract aesthetic features and find the best cropping window among candidate views. According to Table \ref{tab:fcdb}, the proposed model achieves the best IoU and Disp scores compared to the others.

\subsubsection{FLMS Dataset}

We also evaluate on FLMS dataset and the results are shown in Table \ref{tab:flms}. Following \cite{Li2018A2}, we choose Top-1 maximum IoU (Max IoU) as metric to represent cropping accuracy. In addition to ranking-based methods, two classification-based methods are also compared on FLMS dataset. Fang \emph{et al.} \cite{fang2014automatic} learns an aesthetic-based cropping model by discriminative classifier training. In \cite{wang2017deep}, attention box prediction (ABP) network and aesthetics assessment (AA) network are proposed to model the photo assessment problem as aesthetic quality classification. From experiments in Table \ref{tab:flms}, we can see that our model outperforms other methods in cropping accuracy.
\begin{table}[!tb]
\centering
\begin{tabular}{c|c}
\hline
Methods                              & Top-1 Max IoU \\
\hline
Fang et al. \cite{fang2014automatic} & 0.6998        \\
VFN+SW \cite{Chen2017Learning}       & 0.7265        \\
ABP+AA \cite{wang2017deep}           & 0.8100        \\
A2-RL \cite{Li2018A2}                & 0.8204        \\
VPN \cite{wei2018good}               & 0.8233        \\
\hline
\textbf{LVRN (Ours)}                        & \textbf{0.8434}\\
\hline
\end{tabular}
\caption{Quantitative evaluation on FLMS dataset.}
\label{tab:flms}
\end{table}

\begin{table}[!tb]
\centering
\begin{tabular}{c|c|c|c}
\hline
Methods    & Avg. Candidate       & Avg. IoU & Avg. FPS \\
\hline
VFN+SW     & 137   &0.6328& 0.77 \\
VFN+SW+    & 500   &0.6395& 0.22 \\
VFN+SW++   & 1125  &0.6442& 0.10 \\\hline
\multirow{3}{*}{\textbf{LVRN (Ours)}}       & 344   &0.6773& 197  \\
& 919   &0.6841& 153  \\
& {\bf 1745}& {\bf 0.7100}  & {\bf 125}       \\
\hline
\end{tabular}
\caption{Candidate number analysis on FCDB dataset.}
\label{tab:cand}
\end{table}

\begin{table}[!tb]
\centering
\begin{tabular}{c|c|c|c}
\hline
Methods        & Avg. Candidate   & Avg. IoU       & Avg. FPS        \\
\hline
VFN+SW         & 137  & 0.6328        & 0.77              \\
A2-RL          & 13.56  & 0.6633        & 4.08             \\
VPN            & 895  & 0.6641        & 75             \\
\hline

{\bf LVRN (Ours)}   & {\bf 1745}    & {\bf 0.7100}  & {\bf 125}       \\
\hline
\end{tabular}
\caption{Efficiency evaluation on FCDB dataset.}
\label{tab:time}
\end{table}

\subsubsection{Time Efficiency}
To validate the time efficiency, we compare the time cost between our model and the state-of-the-art methods (VFN, A2-RL and VPN) on FCDB dataset. All the results in Table \ref{tab:time} are evaluate on the same perform with one NVIDIA GeForce 1080 GPU.

The selection of candidate views plays an important role for image cropping. In Table \ref{tab:time}, \emph{Candidate} means the number of bounding boxes used to find the best view (in VFN and VPN) or extract the evaluation feature (in A2-RL). In general, the model using most candidate views in evaluation can most likely find the best results shown in Table \ref{tab:cand}. From Table \ref{tab:time}, the proposed method uses the \textbf{most candidate views} in the \textbf{least time} (120+ frames per second) and achieves the \textbf{best accuracy}.

\subsection{Performance Analysis}

\subsubsection{Performance of Listwise Learning}

To illustrate the effectiveness of listwise learning, we design a contrast experiment shown in this section.
As described in Section \ref{sec:method}, we build the VGG16-based networks without RoIRefine using different ranking losses (pairwise and listwise). We train the networks on CPC dataset and compute the rank scores for all candidate views once time. Following \cite{Chen2017Learning,wei2018good}, the pairwise loss is defined as
\begin{equation}
L(v_i^1,v_i^2) = max \{0, 1+f(v_i^1)-f(v_i^2)\},
\end{equation}
where $v_i^1$ and $v_i^2$ is two views selected in the same image $d_i$, and $v_i^1$ is preferred more than $v_i^2$. For listwise training, the training setting is the same as the Section \ref{sec:roirefine} except that RoIRefine is not used.

The result of pairwise training heavily depends on pair selection because the samples with various distribution will result training bias. In order to compare as much as detailed, we train three models with different selection methods shown in Table \ref{tab:listwise}. Without pair selection, there are more than 2.6 million pairs in CPC dataset. With simple pair selection, we set a threshold 0.5 to drop the pairs with a minor gap of rank score, and generate 1.3 million pairs. With careful pair selection, we train the model following \cite{wei2018good}. The markedly improvement in Table \ref{tab:listwise} shows that listwise training overcomes the problem of the pairwise training.
\begin{table}[!tb]
\centering
\begin{tabular}{c|c}
\hline
 Ranking Loss                        & Avg. IoU       \\
\hline
Pairwise + w/o selection (260W+)     & 0.5355         \\
Pairwise + simple selection (130W+)  & 0.5568         \\
Pairwise + careful selection         & 0.6080         \\\hline
\textbf{Listwise}                    & \textbf{0.6204}\\
\hline
\end{tabular}
\caption{Performance analysis of listwise learning on FCDB dataset.}
\label{tab:listwise}
\end{table}

\begin{table}[!tb]
\centering
\begin{tabular}{c|c|c}
\hline
RoI Operation          & Avg. IoU(Pairwise) & Avg. IoU(Listwise) \\
\hline
w/o                    & 0.6080             & 0.6204             \\
RoIPool                & 0.6526             & 0.6706             \\
RoIAlign               & 0.6709             & 0.6956             \\
RoIWarp                & 0.6732             & 0.6997             \\
\textbf{RoIRefine}     & \textbf{0.6882}    & \textbf{0.7100}    \\
\hline
\end{tabular}
\caption{Performance analysis of RoIRefine on FCDB dataset.}
\label{tab:roirefine}
\end{table}

\begin{figure*}[htb]
\centering
\includegraphics[width=0.16\linewidth]{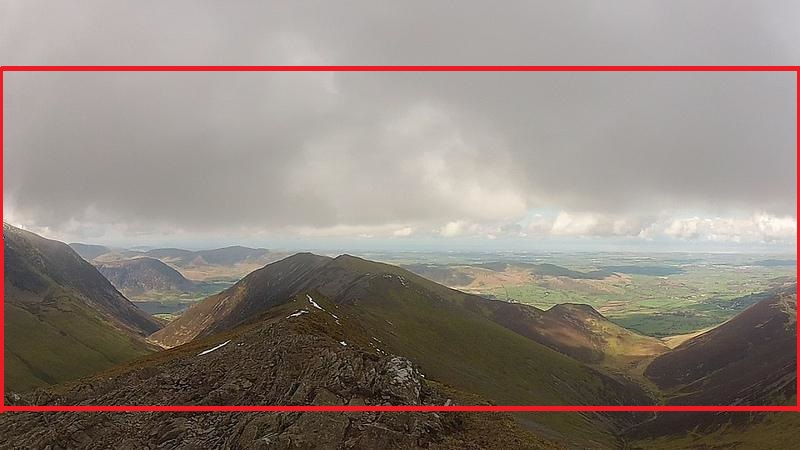}
\includegraphics[width=0.16\linewidth]{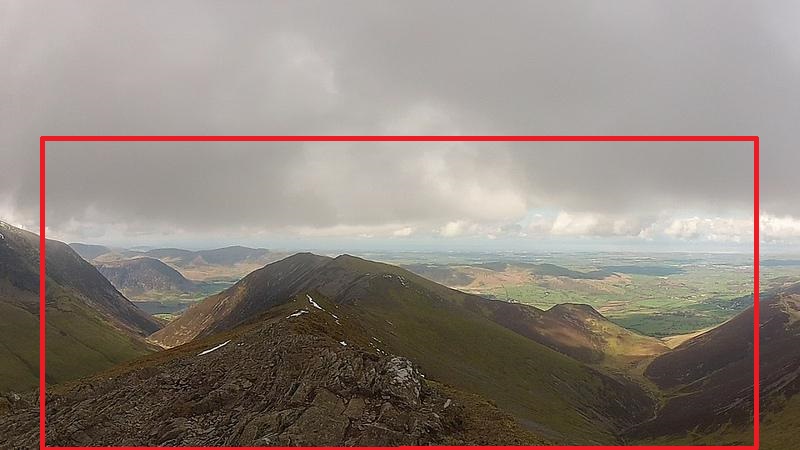}
\includegraphics[width=0.16\linewidth]{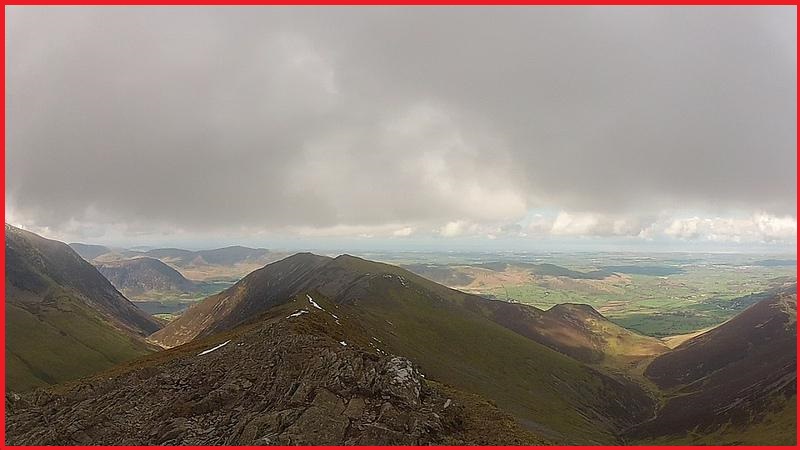}
\includegraphics[width=0.16\linewidth]{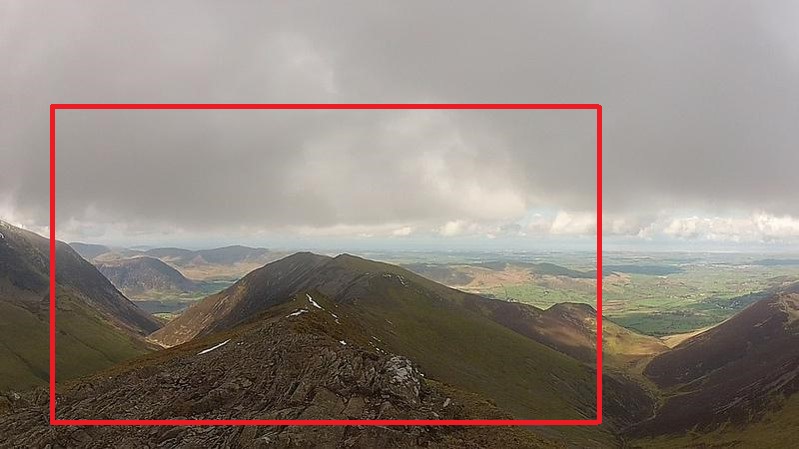}
\includegraphics[width=0.16\linewidth]{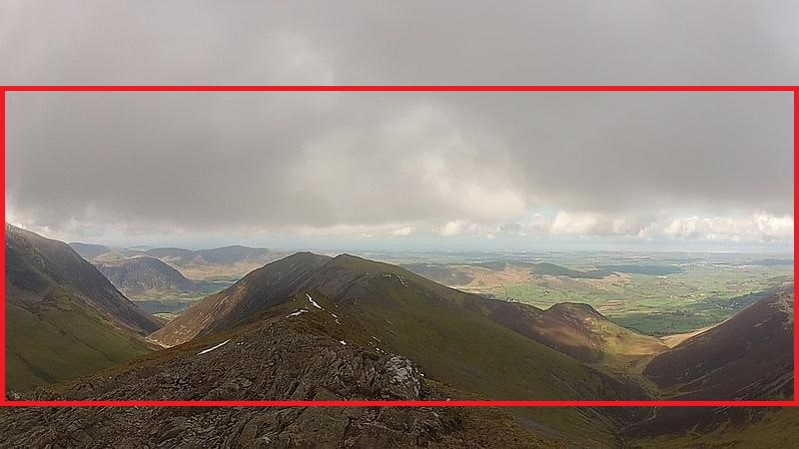}
\includegraphics[width=0.16\linewidth]{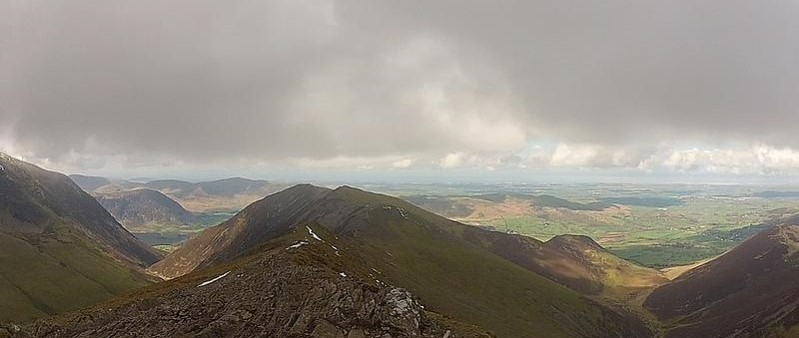}\\[0.1cm]
\includegraphics[width=0.16\linewidth]{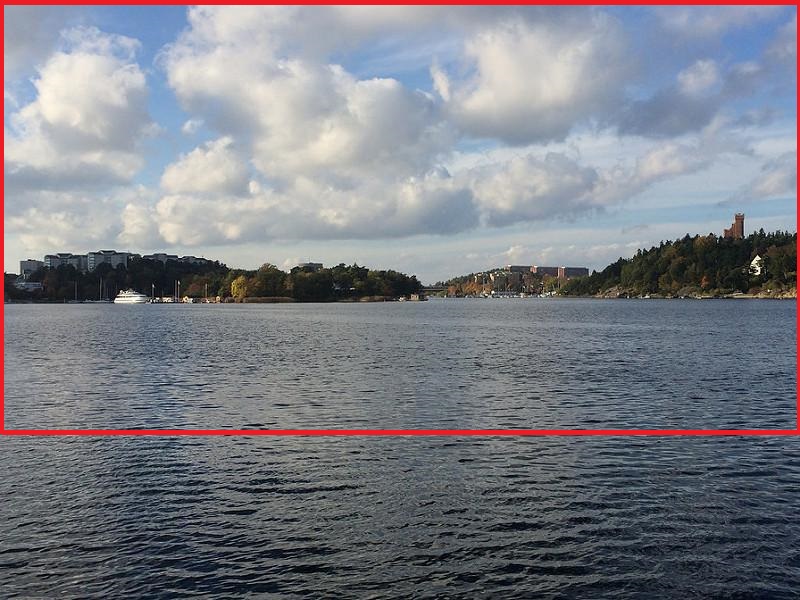}
\includegraphics[width=0.16\linewidth]{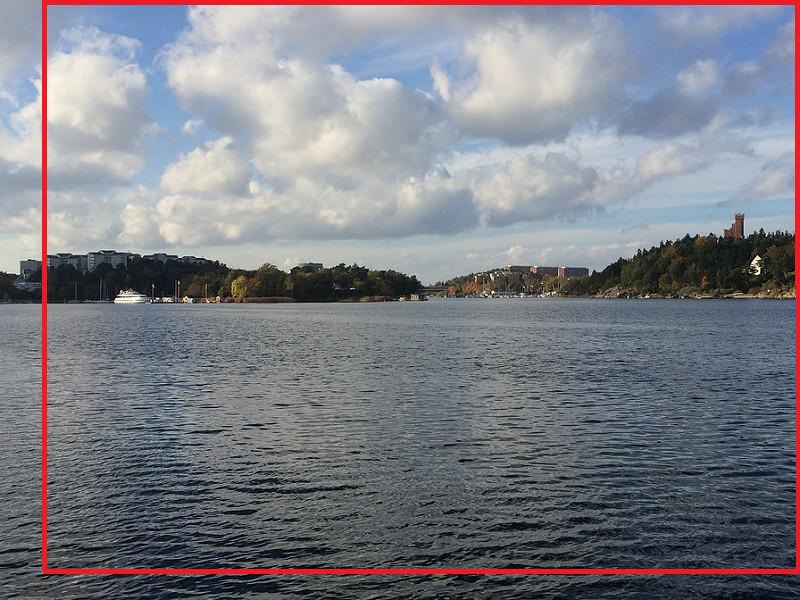}
\includegraphics[width=0.16\linewidth]{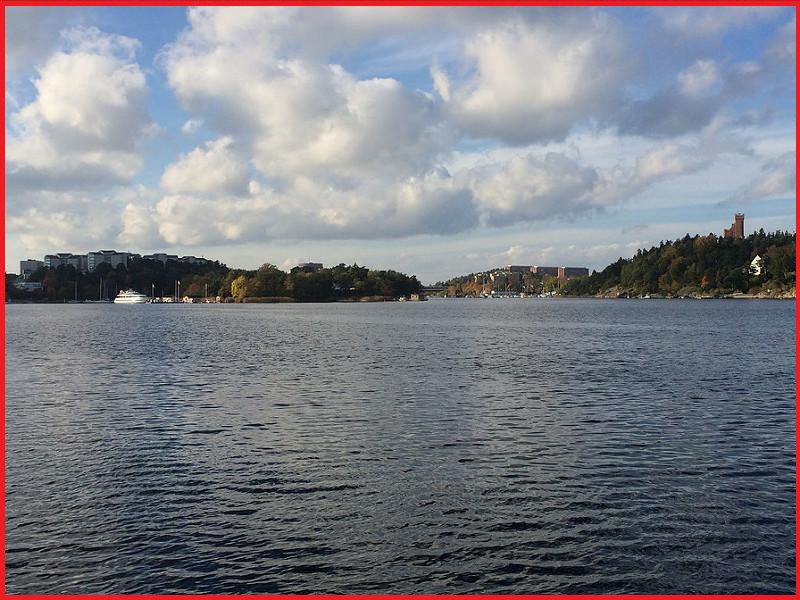}
\includegraphics[width=0.16\linewidth]{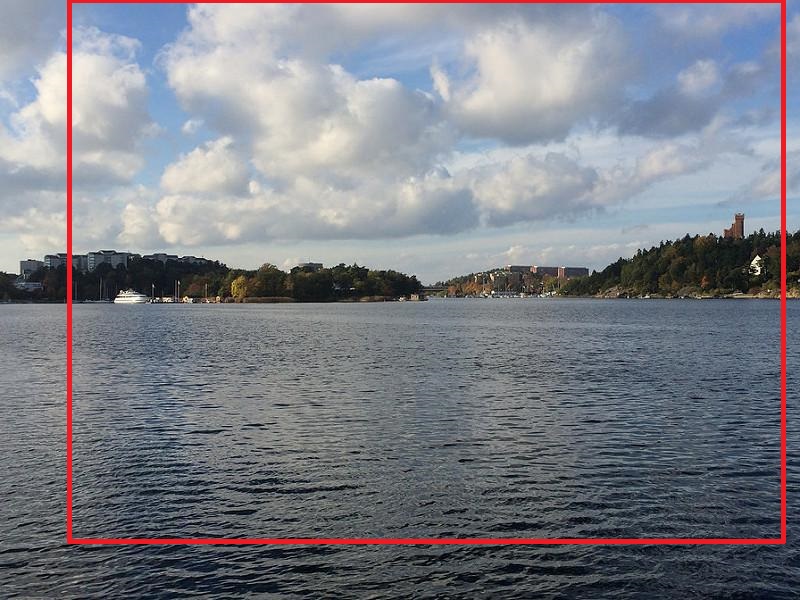}
\includegraphics[width=0.16\linewidth]{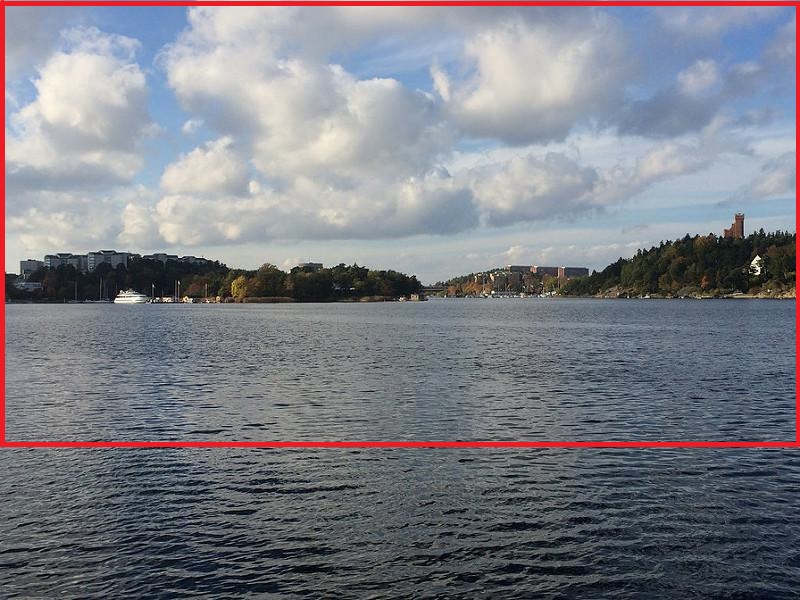}
\includegraphics[width=0.16\linewidth]{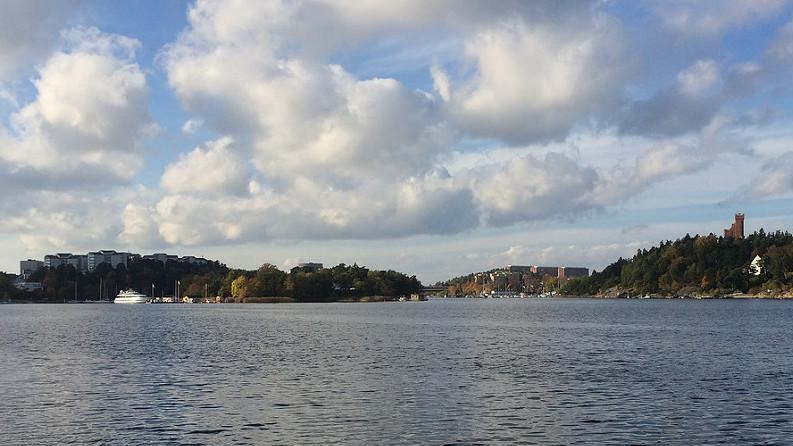}\\[0.1cm]
\includegraphics[width=0.16\linewidth]{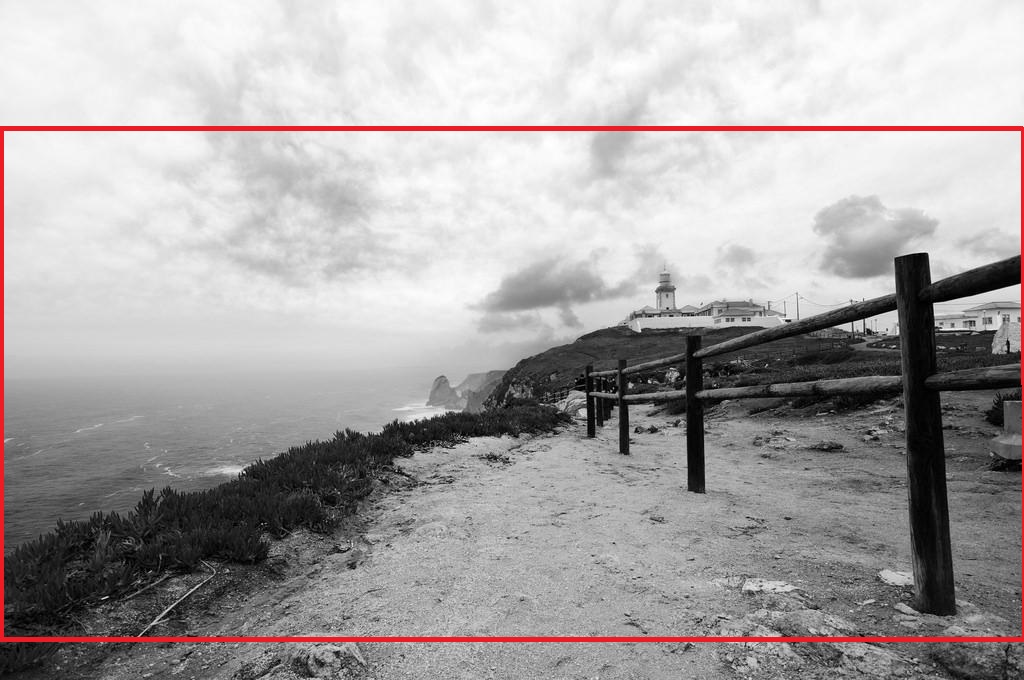}
\includegraphics[width=0.16\linewidth]{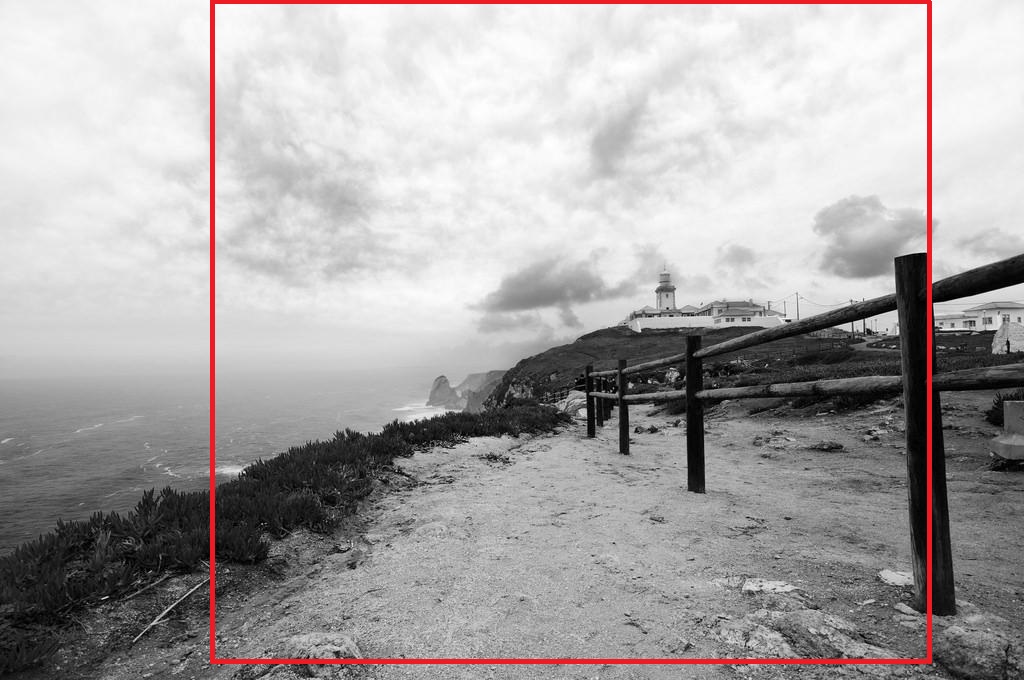}
\includegraphics[width=0.16\linewidth]{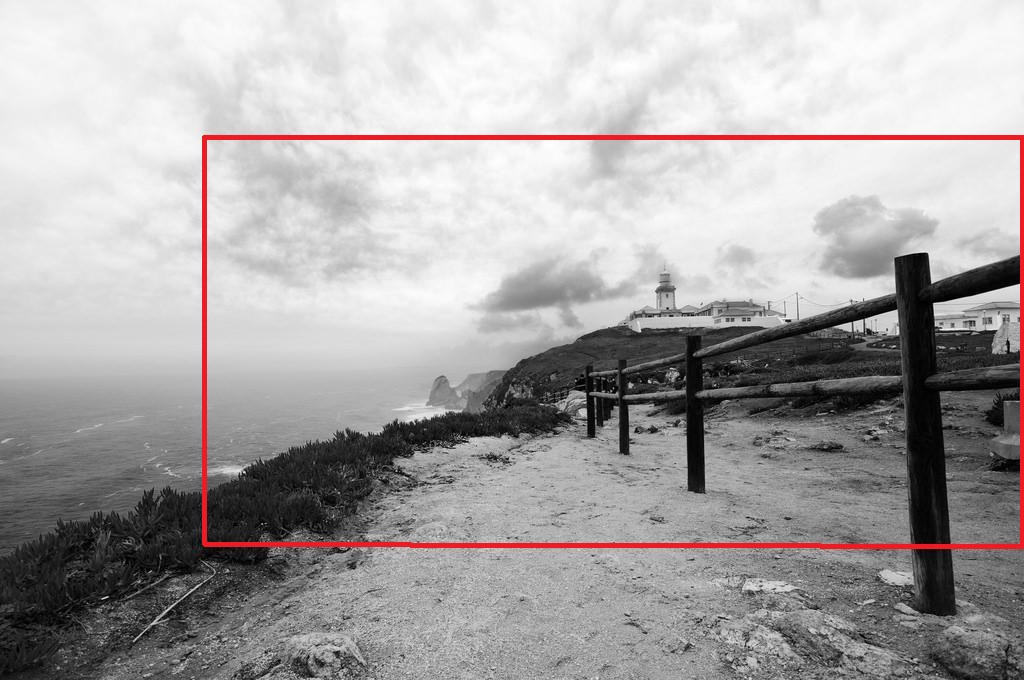}
\includegraphics[width=0.16\linewidth]{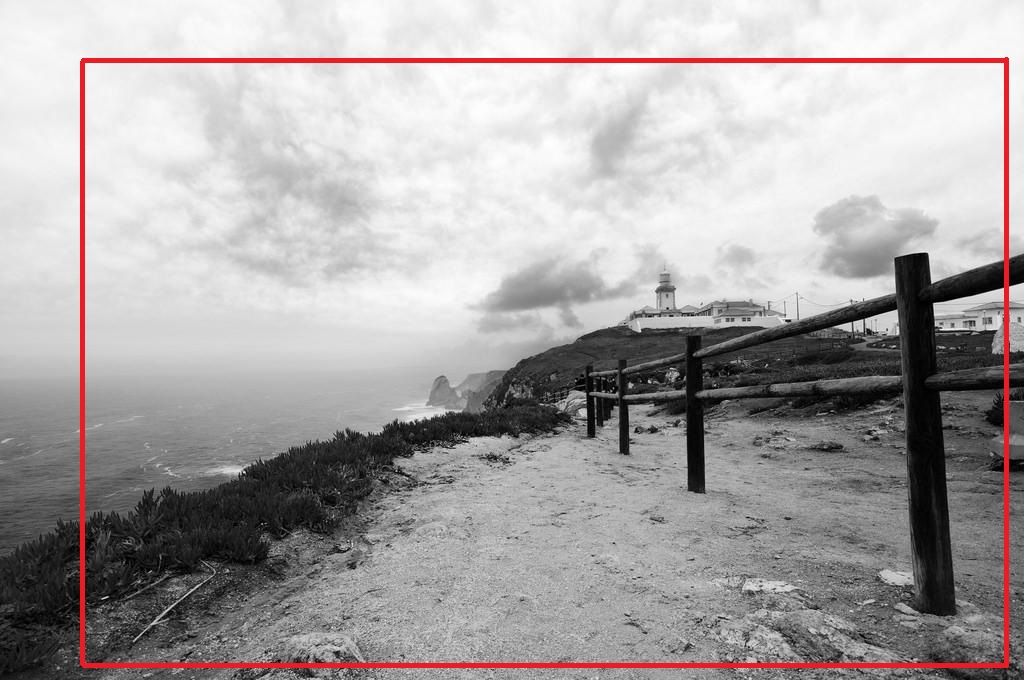}
\includegraphics[width=0.16\linewidth]{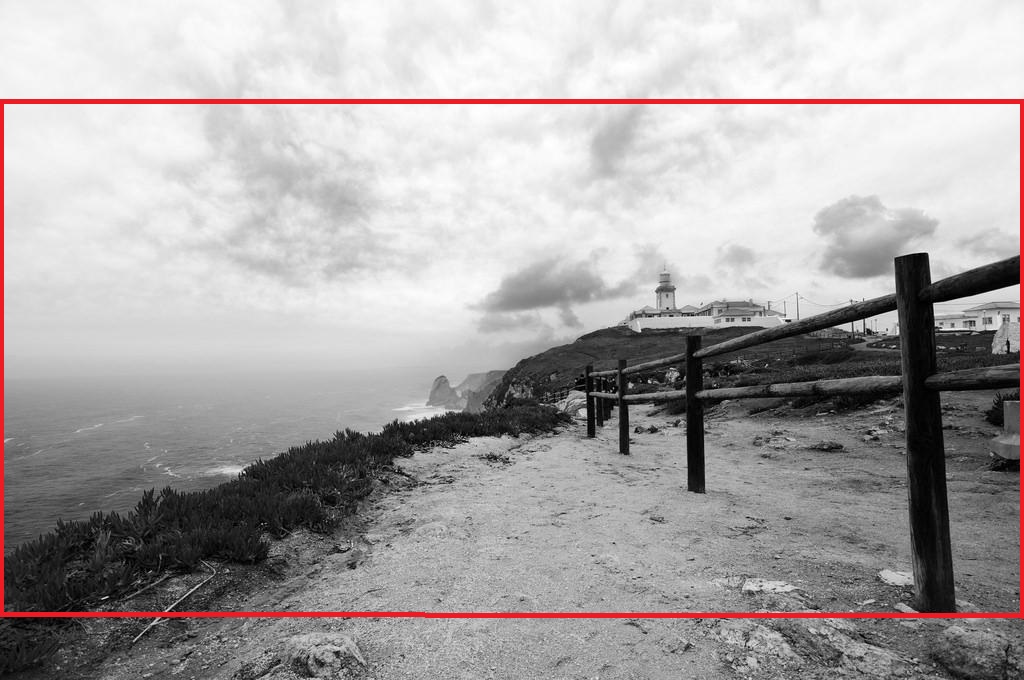}
\includegraphics[width=0.16\linewidth]{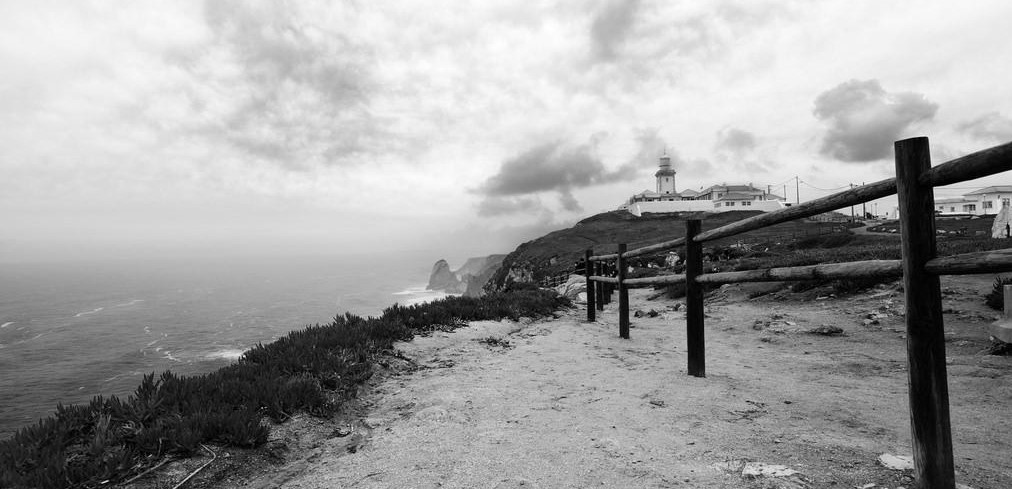}\\[0.1cm]
\includegraphics[width=0.16\linewidth]{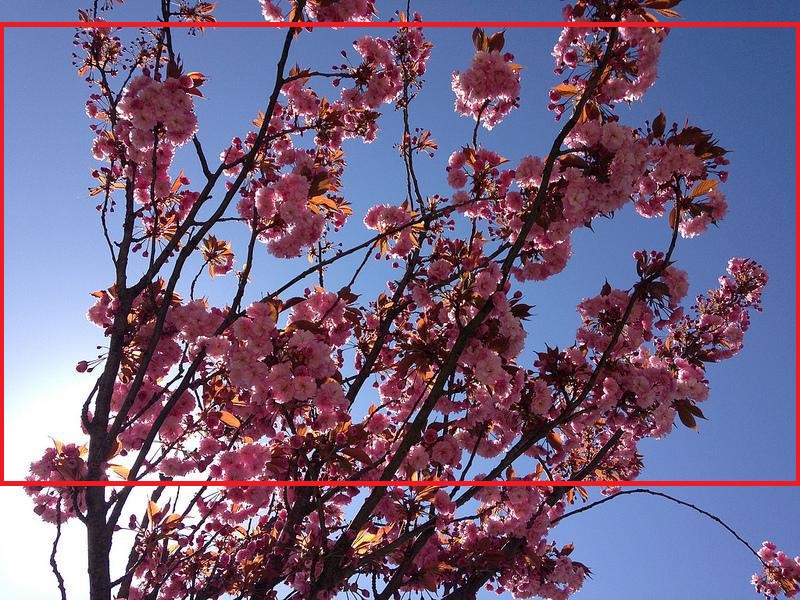}
\includegraphics[width=0.16\linewidth]{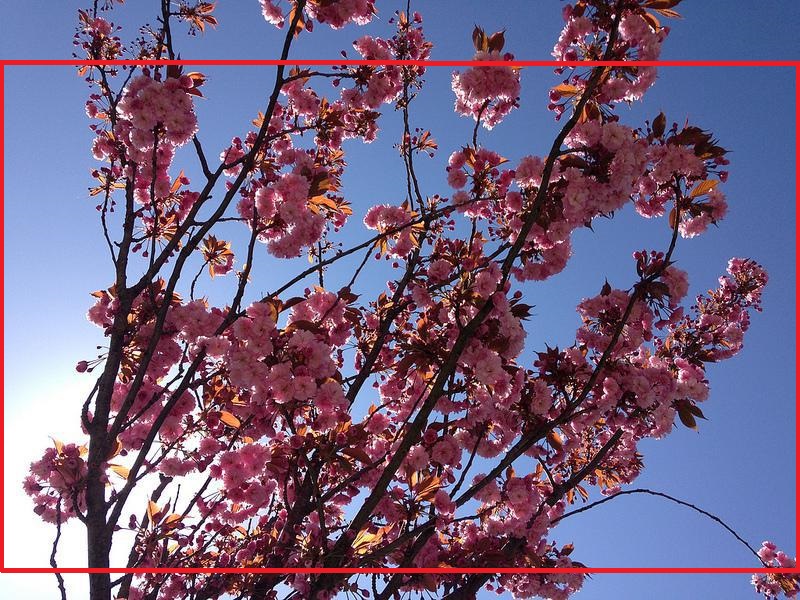}
\includegraphics[width=0.16\linewidth]{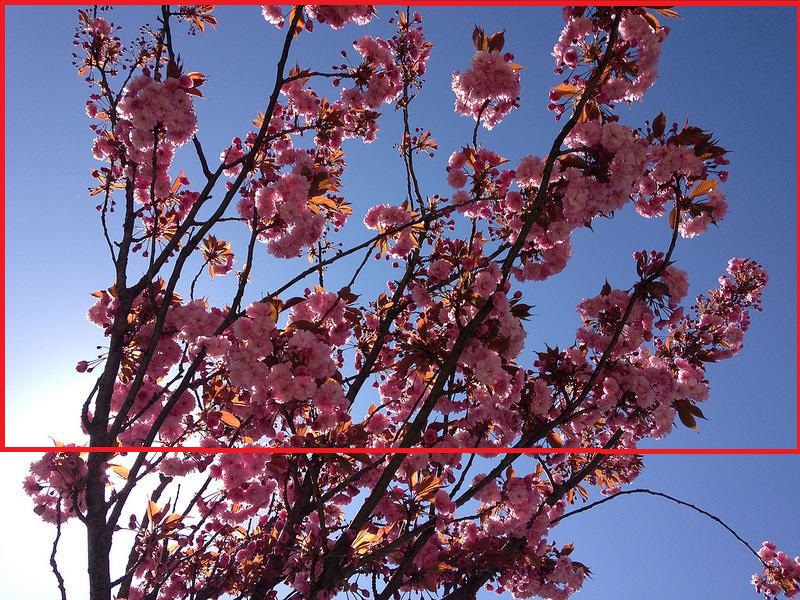}
\includegraphics[width=0.16\linewidth]{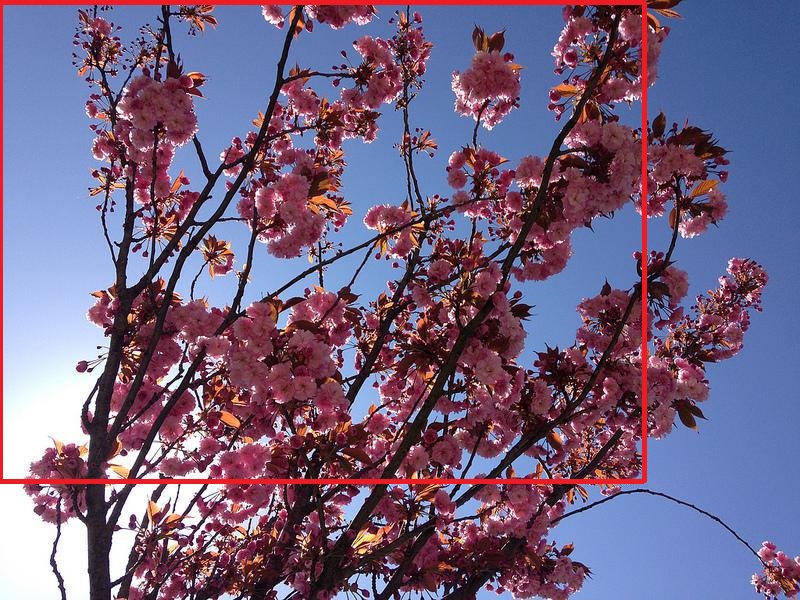}
\includegraphics[width=0.16\linewidth]{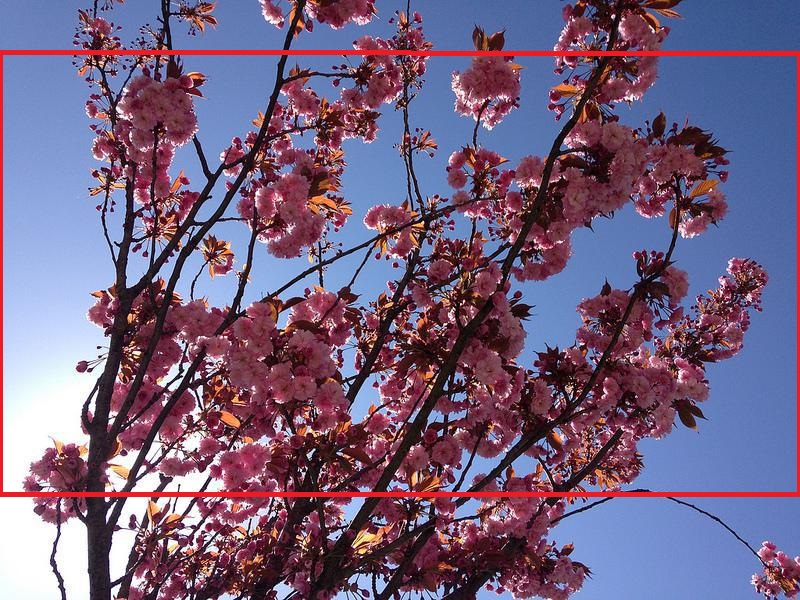}
\includegraphics[width=0.16\linewidth]{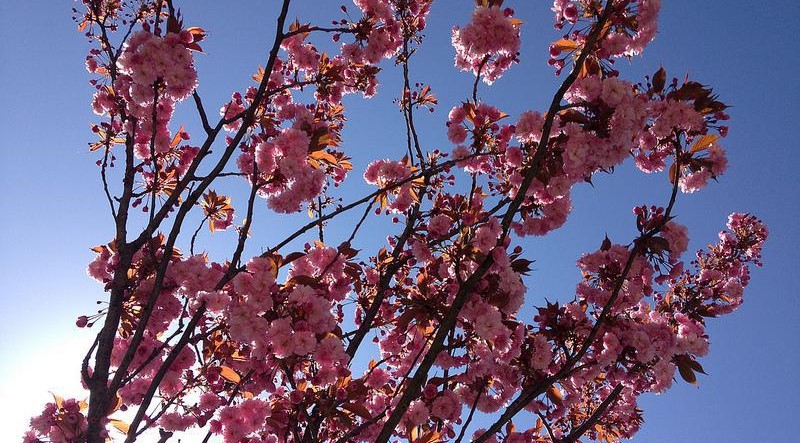}\\[-0.06cm]
\subfigure[Ground-truth]{\label{fig:gt}\includegraphics[width=0.16\linewidth]{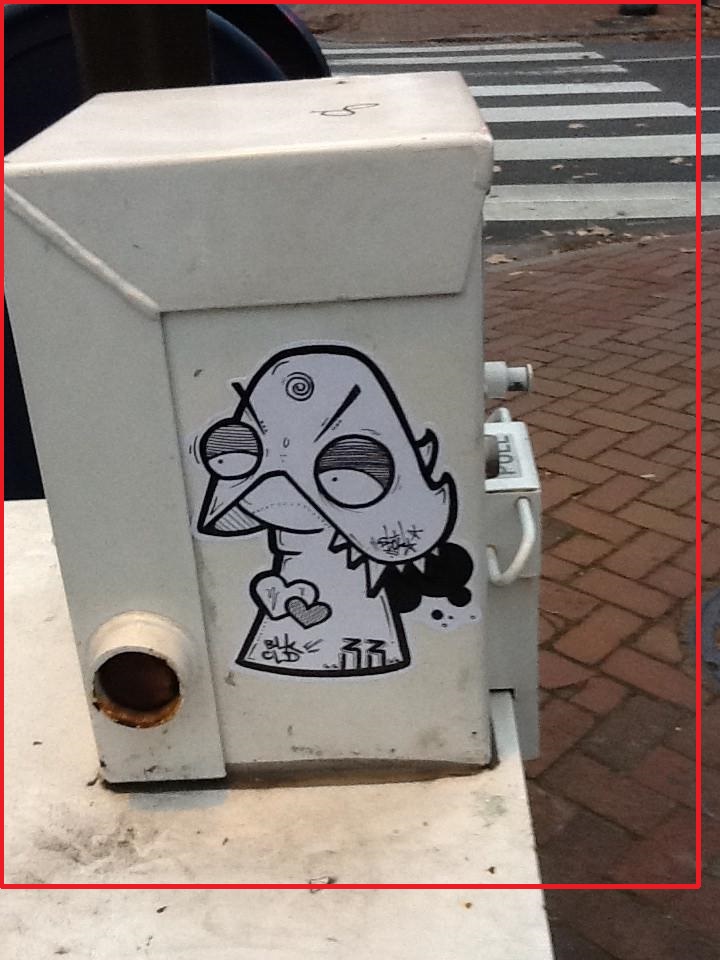}}
\subfigure[A2-RL]{\label{fig:a2rl}\includegraphics[width=0.16\linewidth]{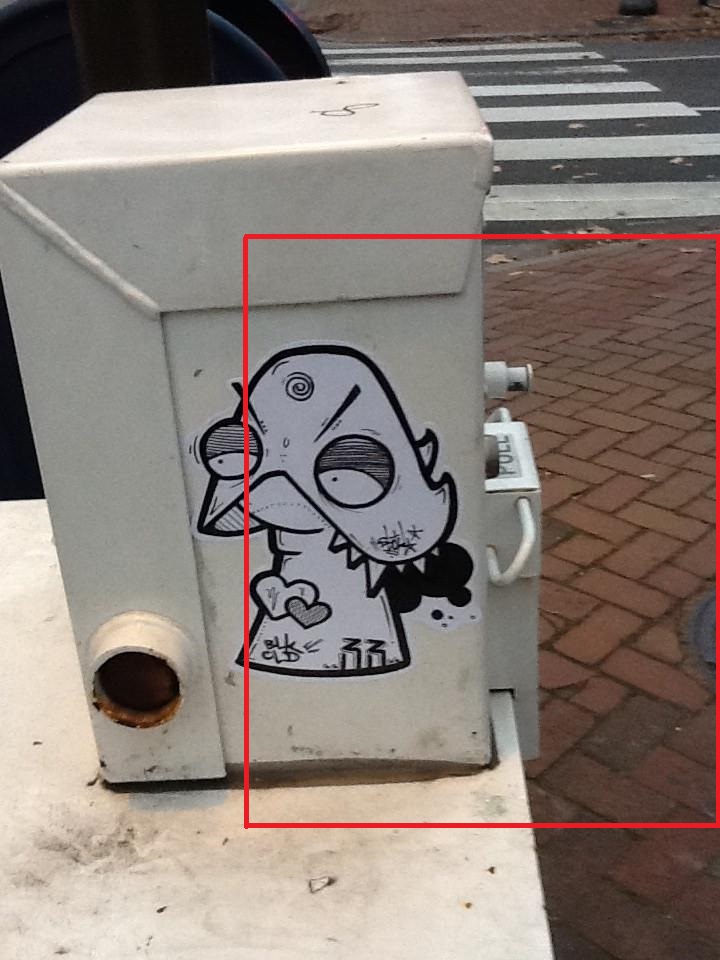}}
\subfigure[VPN]{\label{fig:vpn}\includegraphics[width=0.16\linewidth]{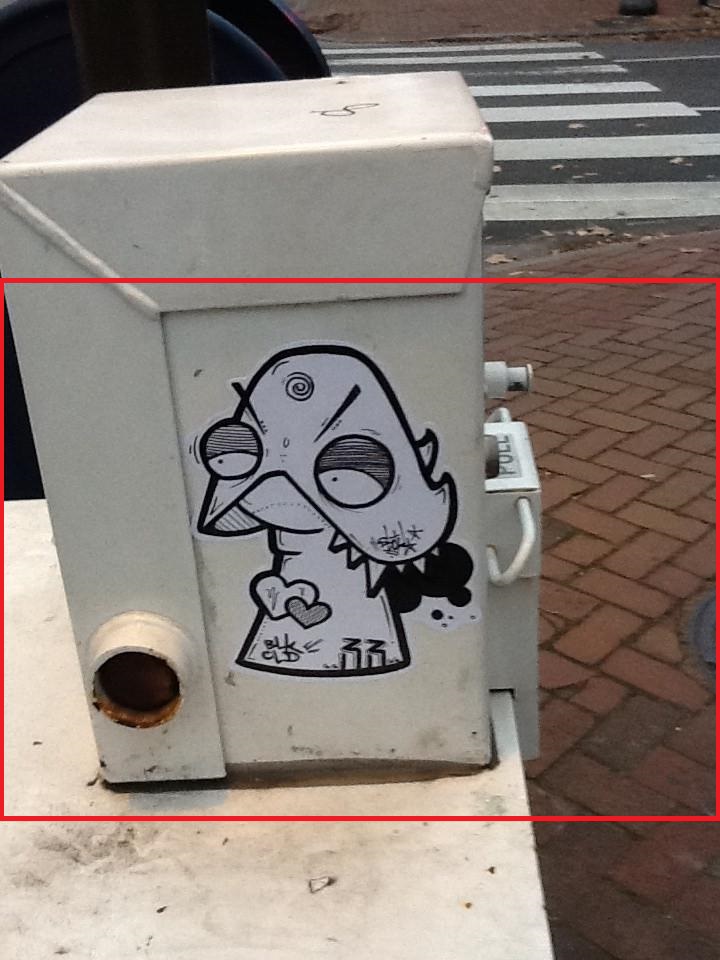}}
\subfigure[VFN]{\label{fig:vfn}\includegraphics[width=0.16\linewidth]{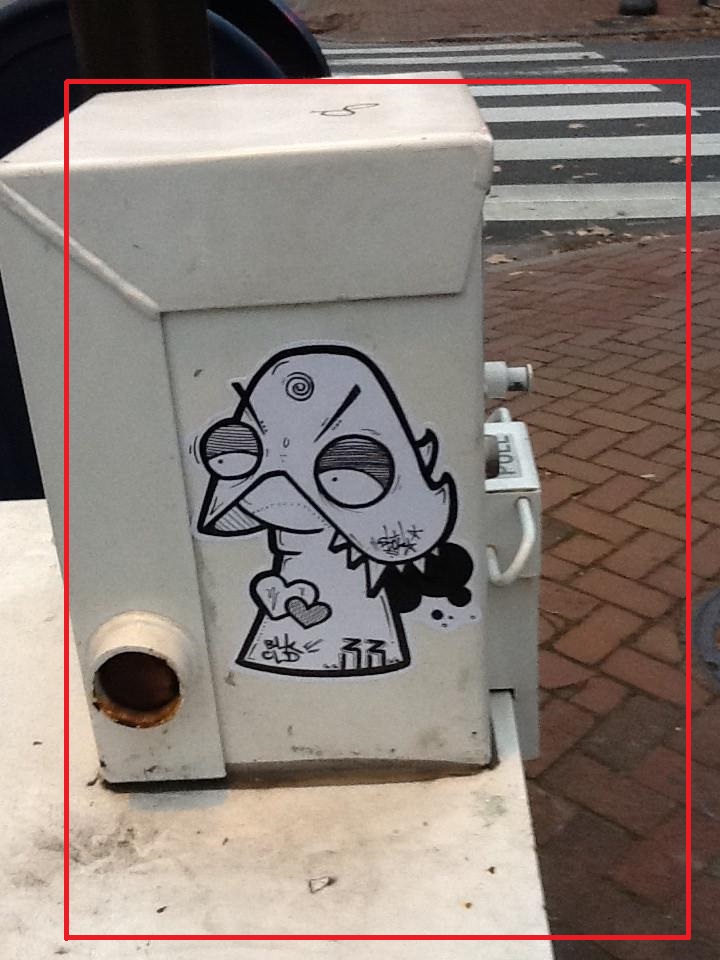}}
\subfigure[LVRN (Ours)]{\label{fig:ours}\includegraphics[width=0.16\linewidth]{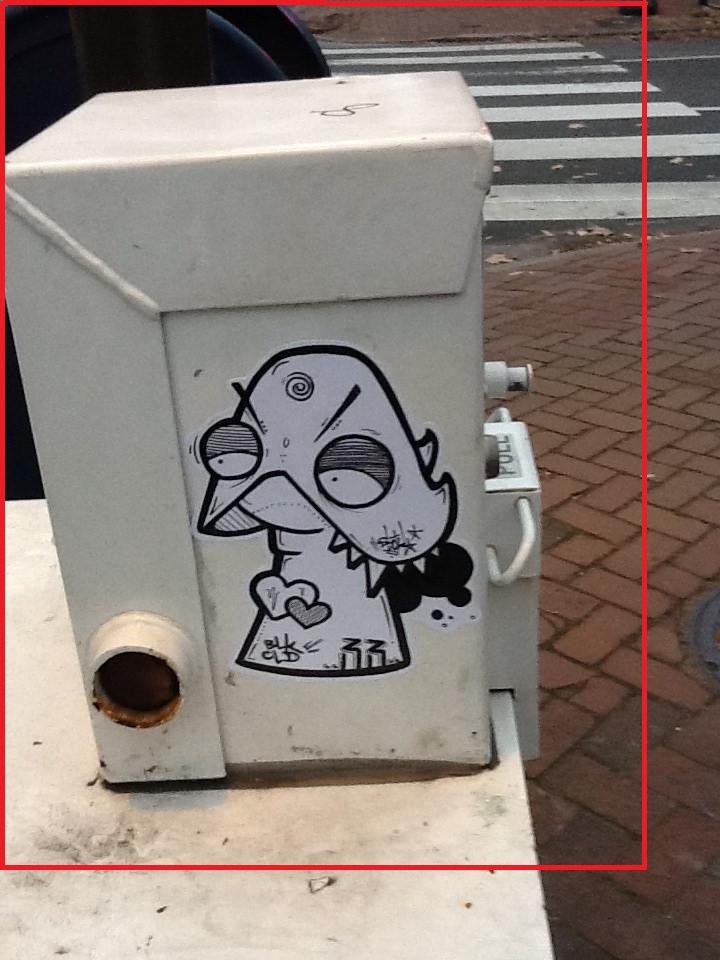}}
\subfigure[Cropping Result]{\label{fig:result}\includegraphics[width=0.16\linewidth]{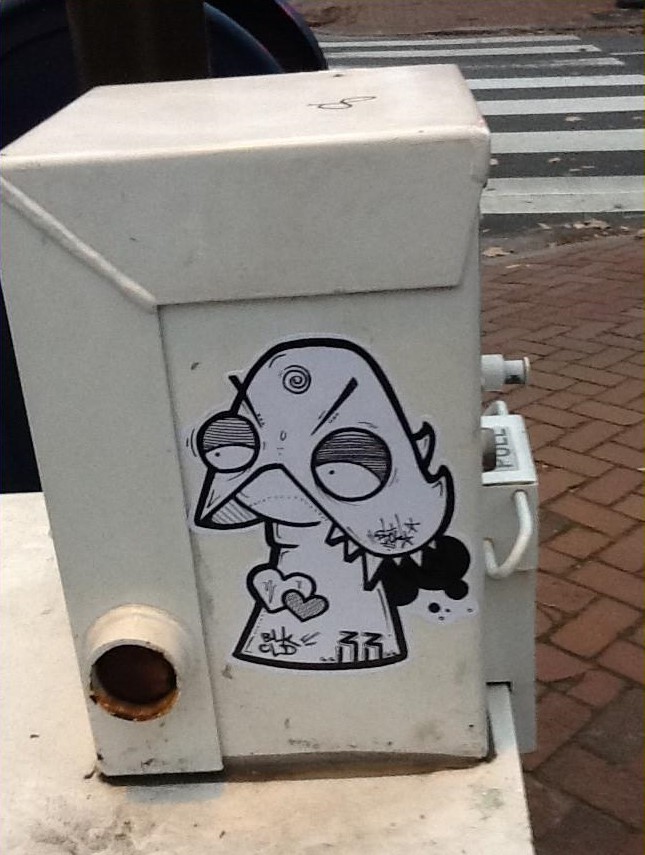}}
\caption{Qualitative visualization on FCDB dataset. In columns (ab) to (e), the best crops determined by different methods are drawn as red rectangles. The column (f) shows the cropping results of our model.}
\label{fig:fcdb}
\end{figure*}

\subsubsection{Performance of RoIRefine}\label{sec:roirefine}
In this section, we study the improvement of the proposed RoIRefine. In order to show the contribution of using the listwise loss function when using the better view sampling(RoIRefine), we train and evaluate eight models using different ranking loss and different RoI-aware operations. The experiment results of RoIRefine and three RoI-aware operations on FCDB are shown in Table \ref{tab:roirefine}. The differences between these RoI-aware operations is the number and place of interpolation shown in Figure \ref{fig:roi}. RoIPool aggregates the view feature after RoI-crop without any interpolation; RoIAlign aligns the feature using interpolation before RoI-crop; RoIWarp resamples the feature using interpolation after RoI-crop. Inspired by RoIAlign and RoIWarp, RoIRefine adopts bilinear interpolation before and after RoI-crop.

RoI-aware operations reduce the deformation caused by traditional view generation and achieve the markedly improvement about 5.0\% IoU. Without interpolation, RoIPool results feature deformation and achieves the worst performance of four RoI-aware operations. RoIAlign and RoIWarp removes the harsh quantization of RoI boundaries, and improve IoU by about 2.0\% to 2.5\% over RoIPool. RoIRefine combines pre-interpolation (RoIAlign) and post-interpolation (RoIWarp) to refine the view feature, and achieves a gain about 1.5\% IoU than RoIAlign and 1.0\% IoU than RoIWarp. The experiment results demonstrate that high-quality features extracted by RoIRefine can overcome the problems of rescaling and deformation.

\subsection{Qualitative Visualization}
As shown in Figure \ref{fig:fcdb}, there are five groups of qualitative results generated by different methods on FCDB dataset.
Obviously, it is very intuitive comparison that our model can extract better view than the others.

For A2-RL (Figure \ref{fig:a2rl}), reinforcement learning is sensitive to initial status and iteration step, resulting unstable performance shown in the second and fifth images.
VPN (Figure \ref{fig:vpn}) uses 895 anchor boxes including the origin image, and tends to select the full image shown in the first two images.
Because of high computational complexity, VFN cannot apply a mount of candidate views to achieve high-accuracy results shown in Figure \ref{fig:vfn}.
Comparing Figure \ref{fig:gt} and Figure \ref{fig:ours}, we can see that our predicted boxes are close to ground-truth.
In the last column, the results cropped by our method have better visual quality than the origin images.


\section{Conclusions}
Image cropping is a common photo manipulation process, which improves the overall composition by removing unwanted regions.
In this paper, we formulate the learning of photo composition as a list-wise ranking problem to overcome the problem of pairwise-based approaches.
Furthermore, a novel RoIRefine operation is proposed to extract high-quality features for view generation.
The experiment results on two common datasets show that our method creates new state-of-the-art results with faster speed of 120+ frames per second.

In the future work, we will study multi-task learning to combine composition evaluation and boxes regression.
Unfortunately, how to design the multi-task loss is a problematic issue.
Inspired of the success of detection framework, RCNN-like \cite{girshick2015fast} or SSD-like \cite{liu2016ssd} method will be our first choice.
\bibliographystyle{named}
\bibliography{ijcai19}

\end{document}